# From a Lossless (~1.5:1) Compression Algorithm for Llama2 7B Weights to Variable Precision, Variable Range, Compressed Numeric Data Types for CNNs and LLMs

Vincenzo Liguori       Ocean Logic Pty Ltd       enzo@ocean-logic.com

**Abstract**
**This paper starts with a simple lossless ~1.5:1 compression algorithm for the weights of the Large Language Model (LLM) Llama2 7B [1] that can be implemented in ~200 LUTs in AMD FPGAs, processing over 800 million bfloat16 numbers per second. This framework is then extended to variable precision, variable range, compressed numerical data types that are a user defined super set of both floats and posits [2]. The paper then discusses a simple hardware implementation of such format based on ANS (Asymmetrical Numeral Systems) [3] that acts as a bridge between this flexible data format and a computational engine while, at the same time, achieving bandwidth reduction. An example of a token factory using weight compression and sharing is also given.**

# 1. Introduction

This paper attempts to address and reconcile two different issues: the existence of multiple numerical data formats (such as int8, bfloat16, fp8, etc., often non optimal for the application and not directly compatible with one another) and the necessity to reduce their bandwidth requirements, especially in the case of power hungry and slow DRAM.

In other words, we would like to be able to support multiple numerical data formats and use a minimal number of bits to represent them while, at the same, not being penalised by the outliers and forced to use a worst-case number of bits to represent them all.

This is particularly important for LLMs that have a huge number of weights that can come in a variety of formats. This is also true, to a lesser extent, for CNNs. Activations are also likely to benefit from such approach. The emphasis here is on inference, but training is also a possibility.

The paper begins with noting that the frequency distribution of Llama2 7B weights is particularly amenable to compression. A simple method, based on entropy coding, is suggested and a comparison is made with gzip and bzip2.

This basic concept is then extended to include different kinds of floating point numbers, variable size integers, posits and more. This effectively allows the creation of custom data types. A variety of examples are given.

This is followed by a discussion about the entropy coder to use: arithmetic coding, Huffman coding and ANS are compared. A simple and fast hardware implementation is discussed with some examples.

Finally, an example of weight compression and sharing is given for a token factory.

# 2. A Simple Observation

Fig.1 Shows the frequency distribution of the weights of the attention matrix Wq0 (Wq layer 0) for Llama2 7B. This is one of the four attention matrices Wq, Wk, Wv and Wo for the first of the 32 layers of Llama2 7B, each of these being approximately 16 million weights.

One pattern that should immediately stand out is the large amount a small weights compared to a few large ones (in absolute value). This pattern indicates that the weights are highly compressible. Intuitively, this is because it is possible to assign a small number of bits to encode very likely values and a larger number of bits to less likely ones, resulting in a smaller average number of bits to encode all the weights. More formally, this process is known as entropy coding.

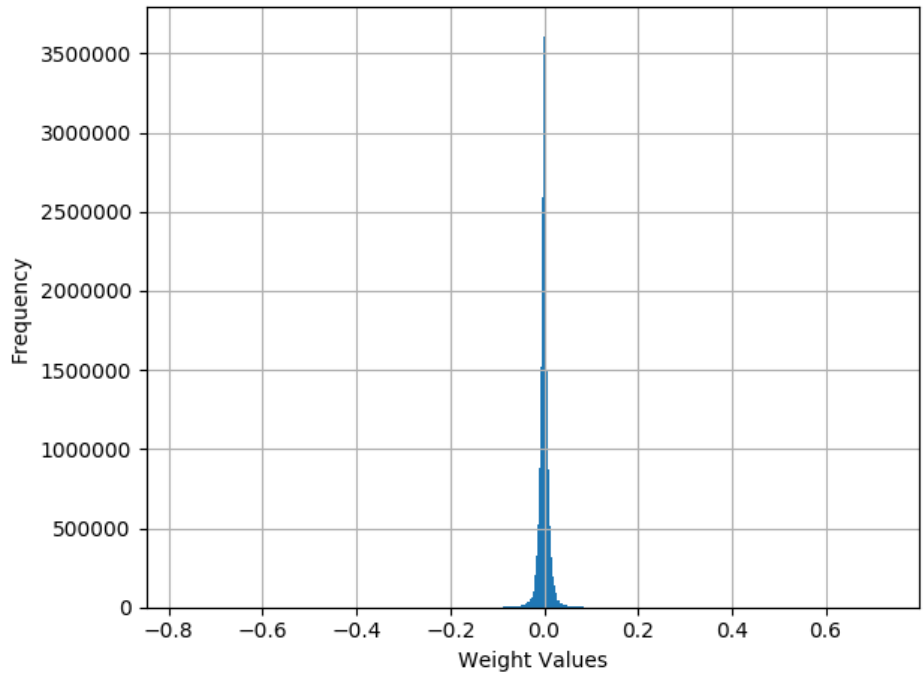

Figure 1  Histogram of frequencies of the weights in the Llama2 7B layer 0 Wq matrix.

This pattern is not a fluke, only to be found in the matrix Wq0: it is repeated for all the other attention matrices, in all 32 layers of Lllama2 7B. This pattern is also present in the weights for all the fully connected layers (the matrices W1, W2 and W3), again in all 32 layers as well as the RMS final weight matrix. In other words, this pattern is present in essentially all the ~7 billion parameters defining Llama2 7B. In this paper I disregarded the other weights as they constitute only a small fraction of the lot.

This pattern is also common in CNNs [4]. For other LLMs, this also seems to be the case [5], see @13:15 of this presentation [6]

We can take advantage of this for lossless compression. Fig.2 a) shows a bfloat16 number. Looking at the distribution in Fig.1, we see that it's pretty symmetrical around zero. This means that there are a similar number of positive and negative values and therefore the information carried by the sign is hardly compressible. Mantissa values are normally uniformly distributed and hence they d o not compress very well. We can set them aside, see Fig.2 b).

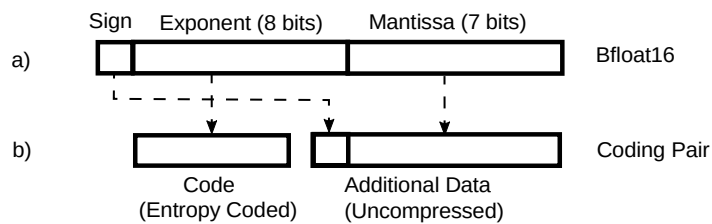

Figure 2 Bfloat16 as a coding pair.

We then notice that not all the possible exponents values are used by the weights: after all Fig1. shows values in a limited range only. We can now count, for each of the weight matrices separately, how many unique exponent values exist. It turns out that for all the matrices but one, there are 31 unique exponent values, 33 for the odd one. This means that we could use a 5 to 6 bits code to encode all the exponents (say 5 for simplicity, shown on Tab.1 under "Simple").

So, looking at Fig.2 b, with this combination of code + additional data (described as "coding pair" from now on) we can losslessly encode any bfloat16 value of the weights with 13 to 14 bits. This translates to savings of some ~2.6 GB from the weights for very little effort .

However, we can do better by using entropy coding on the code value, leaving the additional data uncompressed. In fact, the code values in Fig.2b are not, themselves, equally probable.

## 2.1. Test and Comparison to gzip and bzip2

In order to test the above, I did the following :

- Downloaded Llama2 7B weights from Meta[7], with their permission
- Weights come as fp32, stripped the padding zeros to revert them to bfloat16
- As previously mentioned, some values were stripped from the weights, the resulting, uncompressed, file was 13,214,154,752 bytes
- Each of the weights in the file were converted into coding pairs, as described above
- All coding pairs were compressed with tANS and rANS, two different ANS variants using a customised probability model for each matrix (see below)
- For comparison, compressed the same file with gzip -9 and bzip2 -9

Results are on Tab.1, including a compression estimate for an ideal entropy coder.

| Method | File size (bytes) | % of original size | Avg weight (bits) | Avg code (bits) |
|---|---|---|---|---|
| Original | 13,214,154,752 | 100 | 16 | - |
| Simple | 10,736,500,736 | ~81.25 | 13 | 5 |
| Ideal | 8,735,136,345 | ~66.10 | ~10.58 | ~2.58 |
| rANS 16 bits | 8,738,459,578 | ~66.13 | ~10.58 | ~2.58 |
| tANS 8 bits | 8,826,478,939 | ~66.8 | ~10.69 | ~2.69 |
| Gzip -9 | 10,477,008,576 | ~79.29 | ~12.69 | - |
| Bzip2 -9 | 9,168,474,552 | ~69.38 | ~11.10 | - |

Table 1 A comparison with gzip and bzip2.

In the ideal case we know that, given a code with probability p, it can be encoded with $-\log_2(p)$ bits. We can estimate the probability of a code from its frequency counts and we can then use them to calculate the average number of bits needed to encode a code in the coding pair. When added to the additional data (fixed to 8 bits in this case, see Fig.2), we obtain the average number of bits per coding pair which is the average number of bits used to code a weight in the ideal case. All estimations for the ideal case are performed in double precision in C.

The same estimated probabilities were used to encode the coding pairs with tANS and rANS, with said probabilities reduced to 8 and 16 bits respectively. More details about the ANS hardware implementation are in Section  4..

Gzip and bzip2 are not based on coding pairs (not int the way described here anyway) so code size is absent from Tab.1

The ANS based implementation outperforms both gzip and bzip2 in all cases. This is particularly interesting as the tANS and rANS compressor/decompressors used here have a footprint of ~200 LUTs in AMD FPGAs (for tANS) and are capable of processing a coding pair every clock cycle at 800+ MHz (again tANS).

In comparison, Gzip hardware implementations require one to two order of magnitude more resources, especially for compressing. Not to mention the large memory buffer (I believe 32 Kbytes).

Of course gzip and bzip2 are general compression algorithms while the one presented here is a very specific one. Nevertheless the difference in resource requirement is staggering and it allows, given the same resources, for a very large number of compressors/decompressors in parallel with a very high throughput (see section 4.6.).

More will be said about the ANS implementation later on, but, from now on, we can assume that a simple hardware solution to compress and decompress coding pairs exists. In any case, the discussion here is completely independent from the compression algorithm used.

# 3. Coding Pairs as Custom, Flexible Numerical Data Types

This section contains multiple examples that show the flexibility of the coding pairs working as user defined,variable range, variable precision, compressed numerical data formats.

One important point to remember is that, in a coding pair, the number of bits in the additional data does not need to be the same for each code: to each code value can be associated any number of bits of additional data, including none at all. In other words, the additional data part can be variable in size, depending on the code and different for every code. This gives a lot of extra flexibility, as outlined in the examples below.

## 3.1. Floating Point

We have seen that we can encode bfloat16 with coding pairs and we can do something similar for any other floating point number such fp16 or fp8, including its variations such as E5M2 (5 bits exponent + 1 sign + 2 mantissa) or E4M3.

In fact it is possible to create custom floating point types because coding pairs will easily represent them all. When doing so, it's probably best to use a large exponent as, this way, we get the best of both worlds: one gets rid of "range anxiety" while knowing that the compressor will optimally take care of the outliers without a fixed, worst case size cost.

This means, for example, that if we are using fp8 E5M2 or E4M3, we should use fp11 E8M2 or fp12 E8M3 instead.

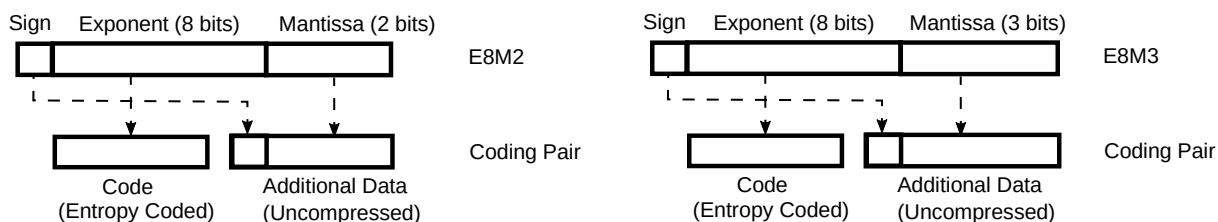

Figure 3: fp11 E8M2 and fp12 E8M3 as coding pairs.

And we can know exactly how well the Llama2 7B weights would compress if they were converted to these two unusual formats (by rounding the mantissa from 7 bits to 2 and 3 respectively). In fact, the exponents are exactly the same as before and so they will compress exactly as in the example above and we can simply reuse the previous result, taking into account the reduced mantissa.

From Tab.1 we know that a code is compressed to an average of ~2.6 bits. So, for fp11 E8M2, we have ~2.6 + 1 bit sign + 2 mantissa = ~5.6 bits/weight. For fp12 E8M3 is ~6.6 bits/weight. Note that these both use less bits and have a much better range than fp8 E5M2 or E4M3.

By the way, this is not just a theoretical discussion: it's practical and serious enough for NVIDIA to include it in their documentation [8] and presentations [9] as well as adding specific hardware to their GPUs.

One more thing: this is not to say that converting Lllama2 7B weights to fp11 and fp12 is a good idea, just how well they would compress if one did.

## 3.2. Posits

I've never used posits, but I loved the video [2]. They are an interesting numerical data type with a variable size exponent and mantissa. In any case, at the end of the day, the regime bits and the exponent bits, together, form an exponent. The remaining bits, if any, form the mantissa. Thus they can be encoded as coding pairs with a variable numbers of bits constituting the additional data and code assigned to occurring exponents.

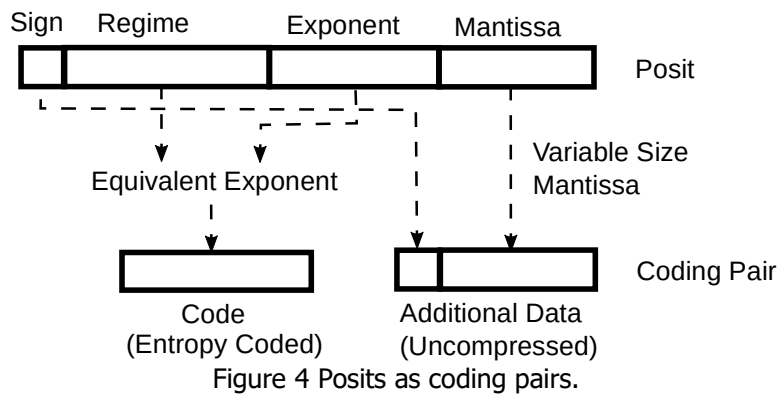

Figure 4 Posits as coding pairs.

I have include them here as an example of the flexibility of the coding pairs with additional data formed by a variable number of bits that depends on the code itself.

## 3.3. Integers

Integer weights are common in CNNs and LLMs weights, generally as a result of quantization,

Unsurprisingly, if we start with a distributions of values such as in Fig.1, then, after a linear quantization procedure such as in [10], we will obtain integers that also have a similar distribution (Fig.5).

And integers with that type of distribution (lots of small value and progressively less large ones) are definitely compressible.

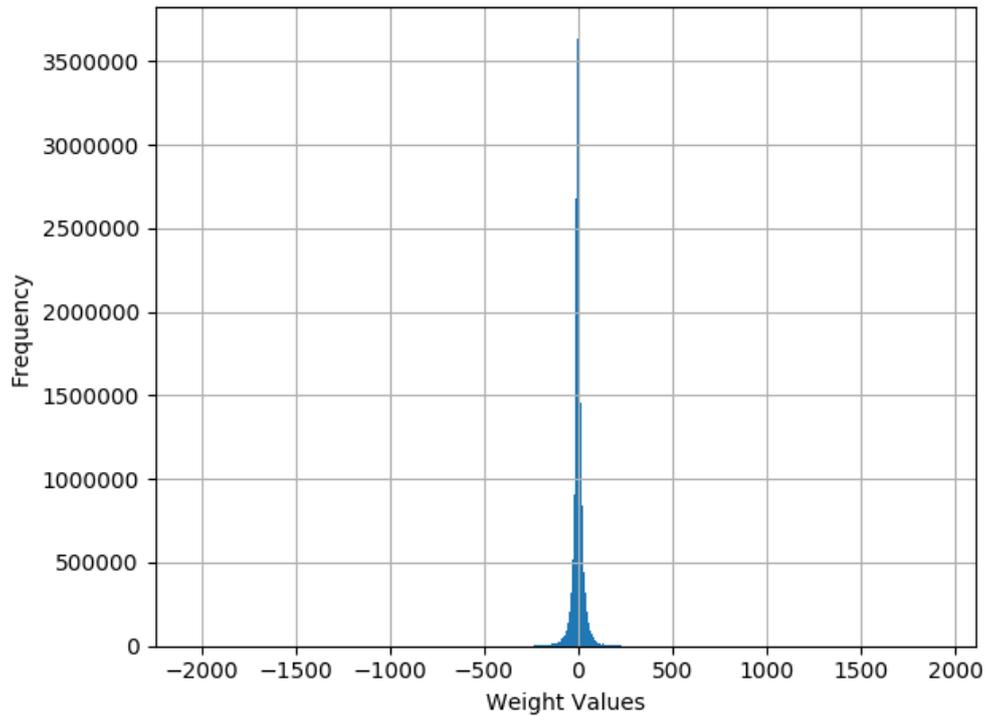

Figure 5 Weight matrix Wq0 quantized to 12 bits ($N_b = 11$).

A possible way of encoding integers with such distribution is the following: zero gets its own code, no additional data. For any other number, the absolute value is taken (sign encoded in the additional data). The position of the non zero MSB (NZ MSB) is found. Such position (plus one to distinguish it from the zero code) now constitutes the code.

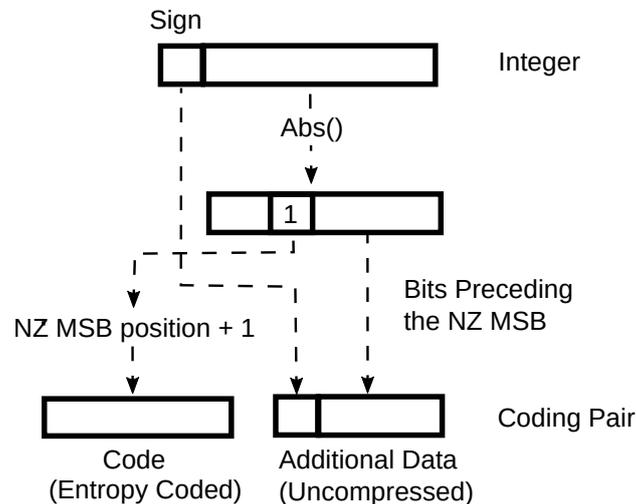

Figure 6 Integers as coding pairs.

The NZ MSB is always one, so it doesn't need to be encoded: the additional data is constituted by the sign plus the other bits. Example : -1, abs(-1)=1, position of NZ MSB is 0, code 0 is already used by the value zero so 0+1=1. Code is 1 and the additional data only contains the sign (1 bit only). Another, 13 = 1101b: NZ MSB position is 3, code is 4, additional data is 101b + sign, 4 bits. Etc. So, for each code k, we have k bits of additional data.

Note that this representation is the same as the floating point representation of the integers but with a variable size mantissa that depends on the exponent. This will make it particularly simple for the interfacing logic proposed in section 4.3.

In this simple coding scheme the additional data size keeps growing with the size of the integer. This precision might be unnecessary for many applications beyond a certain magnitude of the integer (i.e. beyond a certain code) and, at some point, we could limit the size to, say, 8 bits: after all bfloat16 numbers only have 7 bits mantissa. A kind of "saturation".

I have conducted a simple experiment by uniformly quantizing all the same Llama2 7B matrices $W_{ij}$ mentioned in section 2. to $N_b$ bits plus sign, each matrix separately, according to:

$$round\left(\frac{2^{N_b}-1}{max(\|(W_{ij})\|)}W_{ij}\right) \quad \text{Eq.1}$$

I then encoded the resulting integer values according the method above, no "saturation" was used. I then estimated the size of the codes from $-\log_2(p)$ then adding the number of bits of additional data, just as in section 2.1. The result, averaged over all the weights, is in Tab.2.

| $N_b$ | 6 | 7 | 8 | 9 | 10 | 11 |
|---|---|---|---|---|---|---|
| Range | +/- 63 | +/- 127 | +/- 255 | +/- 511 | +/- 1023 | +/- 2047 |
| Avg bits/weight | ~3.391 | ~4.391 | ~5.391 | ~6.392 | ~7.392 | ~8.393 |

Table 2 Average bits/weight after linear quantization and compression.

One final observation: this method of coding integers divides them into exponentially larger regions (i.e. 0, +/- 1, +/- 2-3, +/- 3-7, +/- 8-15, etc.). Within each region, the distribution is assumed to be uniform and thus the additional data that determines a particular integer is left uncompressed. However, looking at Fig.5, it is apparent that this is not the case and this means that better compression is possible.

One possible solution is to subdivide the coding intervals further, especially for intervals where the slope of probability distribution is furthest from the flat. If we bring this process to the limit, it is possible to assign a code for each quantized value: this might be quite practical, especially for lower bit quantization.

For example if we are quantized numbers are 8 bits (7 + sign), then $2^7$ = 128 codes plus the sign as part of the additional data is a perfectly acceptable solution. The hardware that will be introduced later is perfectly capable of supporting 256 or more codes.

In any case, it is apparent, more often than not, that entropy coding of quantized values is definitely beneficial.

## 3.4. Magnitude Outliers

A common problem resulting from LLMs quantization, discussed in [6], is magnitude outliers. In short, as shown in Fig.7 (but, also, in Fig.5), some weights in LLMs are disproportionately large and uncommon compared to the rest.

Now, if we use a fixed number of bits to represent the result of a linear quantization process, we will be forced to chose enough bits for the worst case, used by the uncommon, large weights.

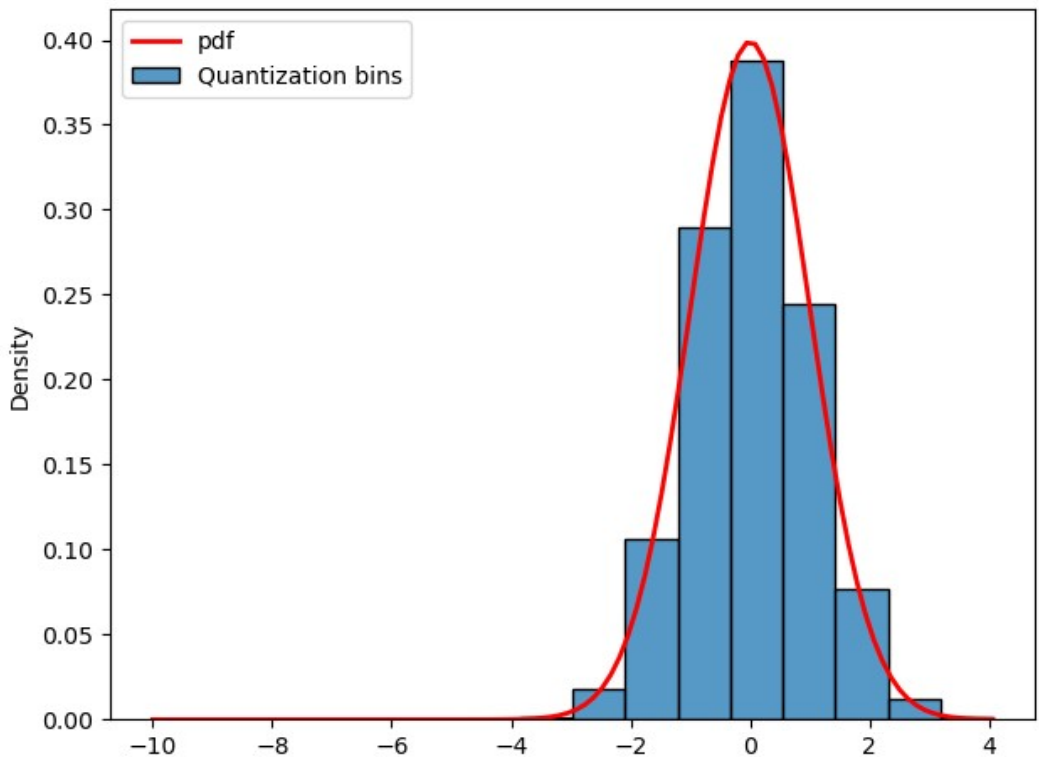

Figure 7: Weight magnitude outliers. Reproduced with permission from [6].

However, if we encode the post quantization values as outlined in the previous section, fewer bits will be assigned to the more common, smaller values with the occasional, larger cost for the larger weights. The overall result will be a smaller representation compared with the number of bits actually used by the quantization, as shown in Tab.2

## 3.5. Weight Saliency: the Other Outliers

Recently there's been a lot quantization work for LLMs such as SpQR[11]. These are based on the realisation that, when it comes to quantization, not all weights are equal. Some are more important than others and should be less heavily quantized (if at all) if performance indicators such as perplexity are to be maintained.

As far as I understand, regardless of the method used to identify an "important weight", once such weights have been found, they are excluded from quantization or subjected to less quantization.

Note that "important" or "ordinary" weights are not necessarily differentiated by their magnitude and they could be, potentially, very similar in value and even have the same exponents.

In previous sections we have associated a code to the exponent (section 3.1.) or a magnitude of an integer (section 3.3.). In fact, a code can be more flexible than that: it can be associated to a particular exponent or magnitude as well as carrying extra information such as whether it is "important" or and ordinary weight. And, of course, different codes can have a different number of bits of additional data, allowing better precision for different codes.

Therefore, thanks to coding pair we can assign different codes to each "important" and "ordinary" and assign different number of bits to the additional data. And we can assign the same exponent two different codes: this way we can encode numbers in the same range in two different ways.

The advantage of doing this is that we do not have to encode "important" weights separately in the bitstream: they can be automatically identified as such and dealt accordingly by the decompressor in the same way shown in section 4.3..

## 3.6. Binary and Ternary Weights

Quantizing to binary and ternary weights has been around for a while, with some recent additions [12], [13]. Here we look at encoding these.

Besides the obvious assigning a code for each 0, +1, -1 (or to 0 and 1 for binary weights), one possibility, if there's a large number of zeros, is to encode the zeros as run lengths plus the sign of the non zero weight at the end. For example, we can encode {0,0,0, -1,1,0,0,1}, from left to right as (3,-1),(0,1)and (2,1).

We can then encode the runs as unsigned integers as in section 3.3. with the sign of the weight included in the additional data. Note that, since the zero runs are positive integers only, there's no need to add their sign to the additional data as in section 3.3.

Of course, if we have binary weights, we only need to encode zero runs as positive integers: since at the end of a run of zeros there's always a 1, no need to add anything.

Encoding binary or ternary weights as run lengths has the additional advantage of being able to skip all the zeros that contribute nothing to a dot product (one of the most important operations for CNNs and LLMs).

This is the approach I used to encode ternary weights for BLMAC (Bit Layer Multiply Accumulator), see [14] or [15]: skip all the zeros and feed the BLMAC with +/-1.

Often, however, grouping weights together could be advantageous from the compression point of view. For example, if we form groups of 8 weights, we might discover that groups of more than, say, 5 non zero weights are unlikely or some other pattern. If there is such or similar pattern an entropy coder will find it and take advantage of it by assigning less bits to more likely patterns.

Fig.8 shows a possible way of doing this by creating a code with the presence of a weight at a certain position. The additional data can then contain just the sign of the non zero weights. For binary weights, obviously, no additional data is necessary. For ternary weights, for example, if we are to encode {0,-1,1,0,0,-1,0,0}, indicating with 1 the position of non zero weights code = 01100100b and the additional data for that particular code will be 3 bits:101b (-1,1,-1).

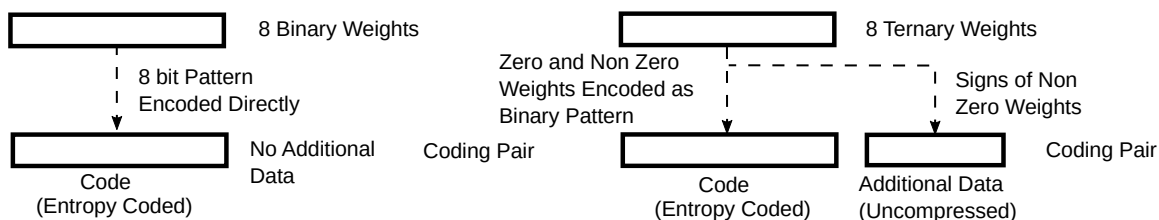

Figure 8 Groups of binary and ternary weights as coding pairs.

If we have lots of zeros, we could create a hybrid system with run lengths of all zero groups with non zero groups encoded as above. This way we could have the best of both worlds: skipping useless zeros and have a compact representation.

## 4. Hardware

So far we have assumed an abstract compressor/decompressor capable of encoding codes inside coding pairs close to -$\log_2(p)$ its, with the additional data coming along for the ride, untouched. Although it was anticipated that the entropy coder would be ANS, nothing that has been described so far relies on a particular entropy coder. We will now see that this is not just a convenient abstraction but it is actually quite simple and practical to **simultaneously support all the formats describe above**.

## 4.1. Choosing an Entropy Coder

As entropy coder, I considered arithmetic coding, Huffman coding and ANS :

**Arithmetic coding**: I have direct implementation experience with CABAC in hardware for a H.264 encoder. While it is adaptive (it can track changing codes' probabilities) and it supports probabilities that are not exact powers of two, it is excruciatingly slow. One has to encode one bit at the time for both the code and the additional data, resulting in tens of cycles per coding pair. I also briefly looked at range coding but it looked more complex than ANS.

**Huffman coding**: I have previously implemented both encoder and decoder in hardware for a JPEG core. Simple encoder but not so simple decoder, at least in the general case. Also, it only supports probabilities for the codes that are integer powers of two. Given the size of LLM models, even an inefficiency of a fraction of a bit can result in a difference of gigabytes in the compressed weight file.

**ANS**: Comes in two flavours: tANS based on a state machine implemented as a table and rANS, similar to range coding but simpler. Both flavours support probabilities that are not exact powers of two and it is possible to compress and decompress a coding pair in a single clock cycle in every case. Main difference from the previous two entropy coders is that it works like a LIFO: if one compresses A, B, C, D, when decompressing, one gets D, C, B, A back. For weight compression during inference, this is irrelevant: if one wants the weights in a given order during decompression, just compress them back to front as this is normally done statically, off line. For activations this is a bit more of a hassle but the amount of data tend to be a lot smaller and so easier to encode in a different order using buffers.

## 4.2. ANS Implementation

I implemented both a tANS and a rANS compressor/decompressor cores capable of processing coding pairs with up to 64 codes and up to 8 bits additional data. The tANS cores support probabilities represented as 8 bits fixed point numbers. The rANS core supports 16 bits probabilities.

Since these cores are extremely useful for a variety of purposes, next step will be creating a generator that creates RTL for different versions with different numbers of codes, different probabilities and additional data sizes.

Characteristics of the cores:

- Guaranteed compression/decompression of a coding pair every clock cycle
- Shallow logic resulting in high clock frequency (6-800 MHz in FPGA, probably 1.5-2 GHz in ASIC)
- Small design size (few hundred LUTs in FPGA)
- Possible to configure number of codes and codes' probabilities. Each code can have a different additional data size associated with it
- Compressor outputs fixed size data words (16 bits for tANS, 32 bits for rANS), as internal bit buffers are filled
- Decompressor requests fixed size data words (16 bits for tANS, 32 bits for rANS), as internal bit buffers are depleted

Note that, by configuring the cores with the number of codes, their associated probabilities and additional data size in bits, it is possible to support **all the coding pairs** described in sections 2. and 3.

Both cores require to be configured with normalised probabilities (defined in Eq 2) for the codes.

$$p_i = n_i / 2^N \text{ with } n_i \in \mathbb{N}, 0 < n_i < 2^N, \sum_{i=0}^{Nc-1} n_i = 2^N \quad \text{Eq. 2}$$

Where N is the number of bits defining the probabilities, Nc is the number of codes. All Eq. 2 is saying is that no code can have zero probability (fair enough: if a code does not exist, the core can't deal with it) and that all probabilities, should add up to 1 (as all well defined probabilities should). Effectively the $n_i$ N bit integers are N bits, purely fractional, fixed point numbers, adding up to 1.

The $n_i$ values are used directly to configure the rANS core whereas, for tANS, they are used to generate state transition tables that are then uploaded to the core.

Normalised probabilities can be generated from code frequency counts with a simple but not trivial algorithm.

Although neither core is adaptive, the fact the rANS core uses normalised probabilities directly opens to the possibility. In fact, one could keep codes frequencies counts up to date while compressing (and decompressing), regularly generating normalised probabilities and update their definition in a rANS core. However, the non triviality of probabilities normalisation would result in a pauses during compression/decompression in order to track/updated the probabilities.

This, in principle, is also possible for the tANS core but with an extra step: after the normalised probabilities are obtained, we need to generate the state machine table before uploading to the core.

As mentioned, tANS is a state machine based on a table. Unfortunately the table size grows exponentially with number of bits of the code probabilities (i.e. 8 bit probabilities => $2^8$ = 256 entries). This makes it less practical for high precision code probabilities. However, its design is simpler than rANS' and, in FPGAs, where distributed memories are an inexpensive resource, it can be a valid option in some cases.

With higher precision probabilities and in ASICs, in general, rANS is possibly a better option.

Tab.3 shows the design implemented in some families of AMD FPGAs with devices with the highest speed rate. The 800+ MHz is reported by Vivado, not sure how practical that is.

|  | Kintex Ultrascale+ | | | Artix Ultrascale+ | | |
| --- | --- | --- | --- | --- | --- | --- |
| Family | LUTs | DSP48 | Max Freq (MHz) | LUTs | DSP48 | Max Freq (MHz) |
| tANS Decoder | 199 | 0 | > 800 | 198 | 0 | > 800 |
| tANS Encoder | 205 | 0 | > 800 | 192 | 0 | > 800 |
| rANS Decoder | 386 | 1 | ~650 | 402 | 1 | ~559 |
| rANS Encoder | 676 | 0 | ~617 | 661 | 0 | ~551 |

Table 3 Implementation figures for AMD FPGAs.

The RTL of the cores described here is targeted to FPGAs but it should be relatively easy to port it to ASIC as it only requires some reduction of distributed memories that are very inexpensive in FPGAs.

## 4.3. Interfacing to a CNN/LLM Processor

This section describes how, with an almost glueless interface, it is possible to input and output from a processor all the compressed numerical data types described in in sections 2. and 3. (with the exception of section 3.6. when binary and ternary weights are dealt as rung lengths or groups). Two cases are given: a floating point processor and a fixed point one.

Suppose that you have designed a LLM/CNN processor supporting internally floating point numbers of sufficient precision to avoid surprises. It might supporting bfloat16 or maybe even fp32 or some other non standard floating point format. You now want to input/output from/to memory all those variable range, variable precision, compressed formats described. This is shown in Fig.9.

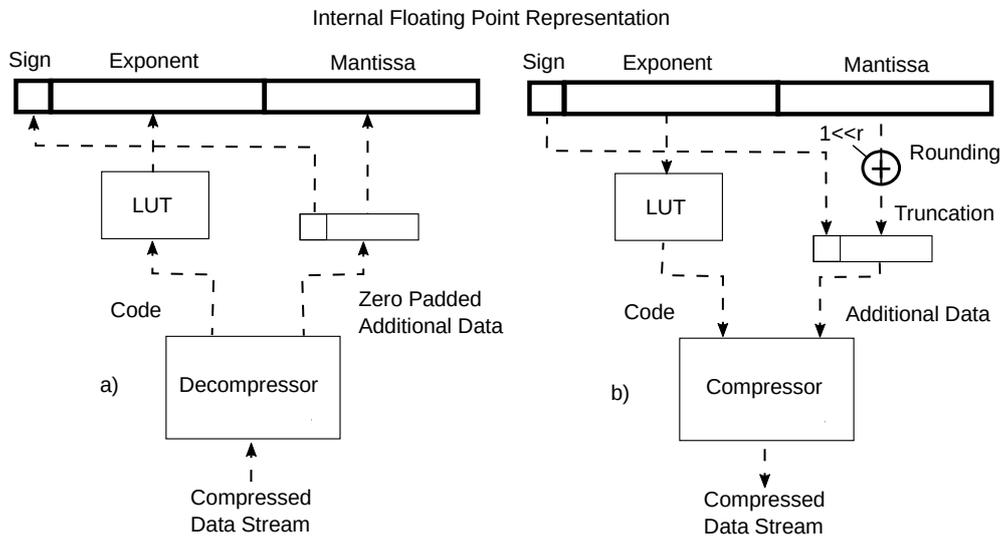

Figure 9 Streaming from compressed data to fp representation.

First we configure the decompressor with the number of codes, probabilities and additional data size information according to the type or coding pairs we want to support. We also configure the look up table (LUT) mentioned below.

Once started, the decompressor (Fig.9 a) will produce a coding pair every clock cycle. All that needs to be done is to map the code to an exponent with a small look up table. The core will even pad with zeros the additional data (to the size of the interface) if the number of bits is too small. Some saturation or underflow logic might be required if exponent of the decompressed numbers does not match the internal representation. A floating point number will be output every clock cycle.

When compressing, we can do something similar: after the compressor (Fig.9 b) is configured, we also initialise the look up table that maps the exponents to a code.

The diagram also shows some rounding logic that rounds the presumably more precise internal representation to the one of the additional data associated to the code. Since, from its configuration, the compressor knows how many additional bits a specific coding pair should contain, it can automatically truncate the extra bits but rounding is necessary beforehand. I have shown the rounding explicitly but, since the compressor knows how many bis are required, its rounding is probably best done inside.

Once the compressor is started, a floating point number can be processed every clock cycle.

For a fixed point processor things are slightly more complicated (Fig.10).

In the decompressor (Fig.10a), the mantissa is converted to a two's complement integer from the additional data. Again, during decompression the code is mapped to an exponent with a look up table. This is then used to shift the reconstructed integer to fit with the fixed point representation.

During compression (Fig.10b) the fixed point value is essentially converted to a float with the exponent mapped to a code with a look at table. This is done by a process called normalisation which, at least principle, can be performed with a priority encoder and a shifter. The discussion about rounding is essentially the same as before. So, once the compressor/decompressor is initialised, it can compress/decompress a number every clock cycle.

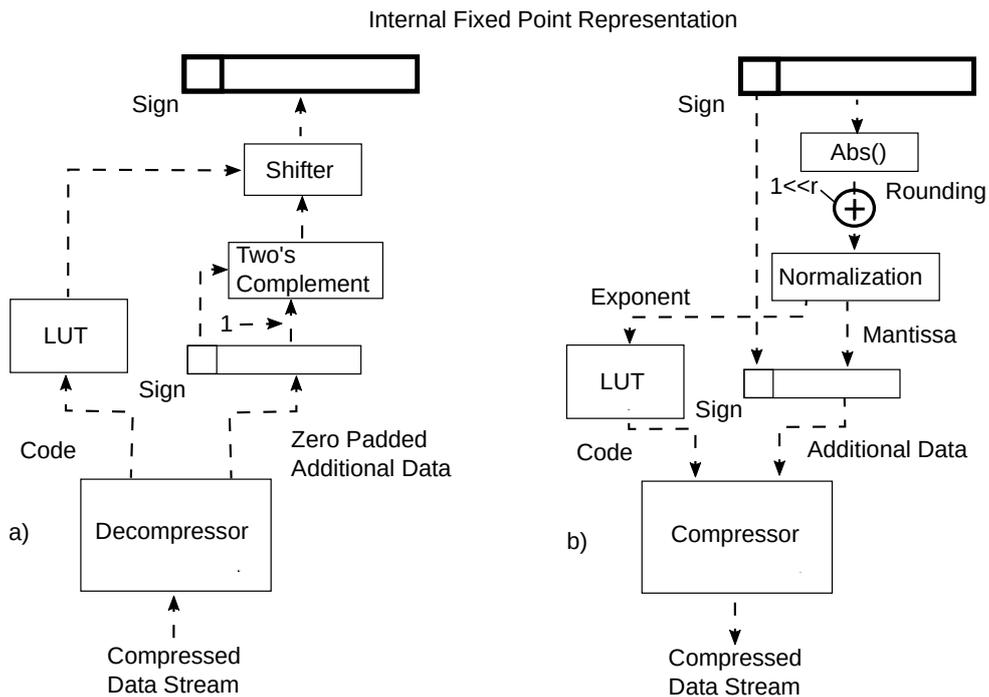
Figure 10 Streaming from compressed data to fixed point representation.

In order to also support the case where a value is mapped directly to a code, as described, for example, at the end of section 3.3., we can have a bigger LUT containing the value associated to each code and use a

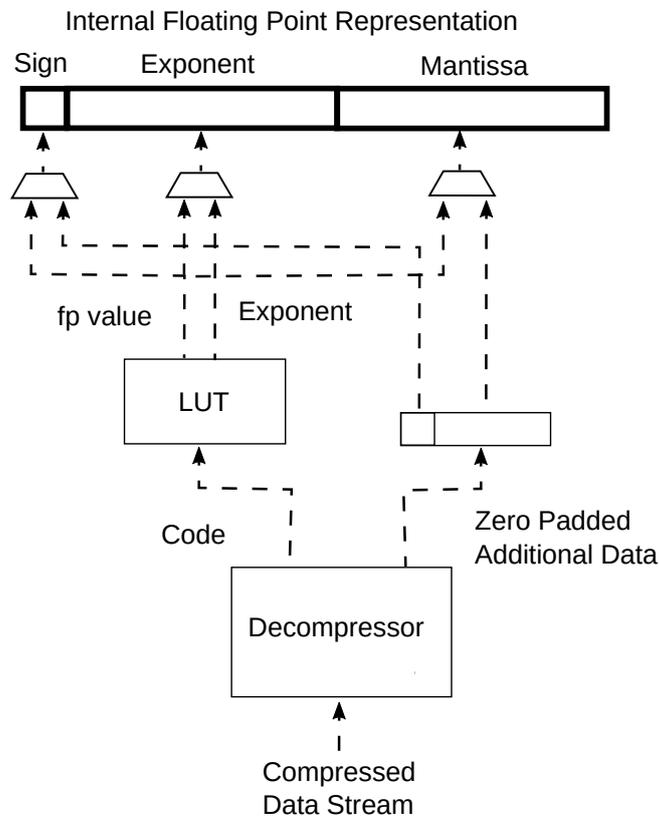
Figure 11 Adding support for direct coding of values

multiplexer to bypass the normal path from additional data and use the value directly. This can also be useful for formats such as fp4 (4 bits floating point). See Fig. 11. The multiplexers can also be controlled from the look up table. Multiplexers' control was not shown in the picture for clarity and it allows to support both methods simultaneously and the code itself (through the LUT) can select which one.

It's easy to see how all the variable range, variable precision, compressed data types mentioned at the start of this section can all be supported by small amounts of glue logic added to the compressors/decompressors. This allows an enormous amount flexibility in different situations for the computational engine.

## 4.4. Random Access to Compressed Data

Access to compressed data is generally sequential: numerical values in various format are streamed from or to a compressed data in memory. This is not really a problem for LLMs and CNNs as, at the heart of both of them, there are large dot product operations that require streaming data.

However, some random access is possible through the equivalent of pointer-like structures even though pointer arithmetic is not.

For example, if we want to access a particular element in an Huffman encoded data in memory we need, obviously, a pointer to its memory location. But we also need a pointer to a bit within the word where the compressed element it located. So, for Huffman coding, to access a compressed element in the data stream, we need a pointer-like structure consisting a pointer to memory as well as a (smaller) bit pointer.

For ANS it is a bit more complicated bur essentially the same: besides the memory location and a bit pointer, we also need some information on the status of the decompressor that is specific to that particular location. Similarly, if we want to append data to compressed data already in memory, we need restore status information in the compressor in order to guarantee that is decompressable when eventually retrieved.

As for the reasons to access random points in a compressed data stream, we might need to access a particular column in a compressed matrix of weights. Or, we might need to read back activations that we have previously stored in compressed format.

## 4.5. Activations

Activations seem to have frequency distributions amenable to compression as well [5], also @27:00 in [16]. However it is harder to exploit because exact frequency counts to extract code probabilities can only be collected at runtime. This is unlike the weights whose probabilities, at least in the case of inference, can be statically determined very precisely.

There are a couple of options. We can, at runtime, for each block of activations produced, collect the code frequencies, produce the normalized probabilities, use them to program a compressor and then finally compress the data. We also need to store the probabilities in order to reverse the process and read the data back. This requires buffering of the activations and extra delays. For this purpose, rANS is probably the best choice as it only needs normalized probabilities. This is unlike, say, Huffman, that from said probabilities then requires a tree generation, making the process even more convoluted.

A better option might be to collect statistical information on the activations by running a large number of typical data through the LLM/CNN in question. We can then use these to program compressors/decompressors without changing them dynamically. This static information will be less precise, but it will probably still be useful and it is certainly a lot easier than the other option. Since are only less precise estimates, maybe the much simpler tANS cores could be used, especially in FPGAs.

Activations also don't require as much bandwidth as the weights and so there is less incentive in doing something about it. Nevertheless the compressor/decompressors discussed here are simple enough that they should be considered.

## 4.6. Data Streaming

We have seen in section 4.3. that compressor/decompressors can stream user defined numerical data types as compressed data.

It follows naturally that, if the bandwidth provided by one unit is not sufficient, multiple ones can be instantiated in parallel, sharing a memory through FIFOs, as shown in Fig.12.

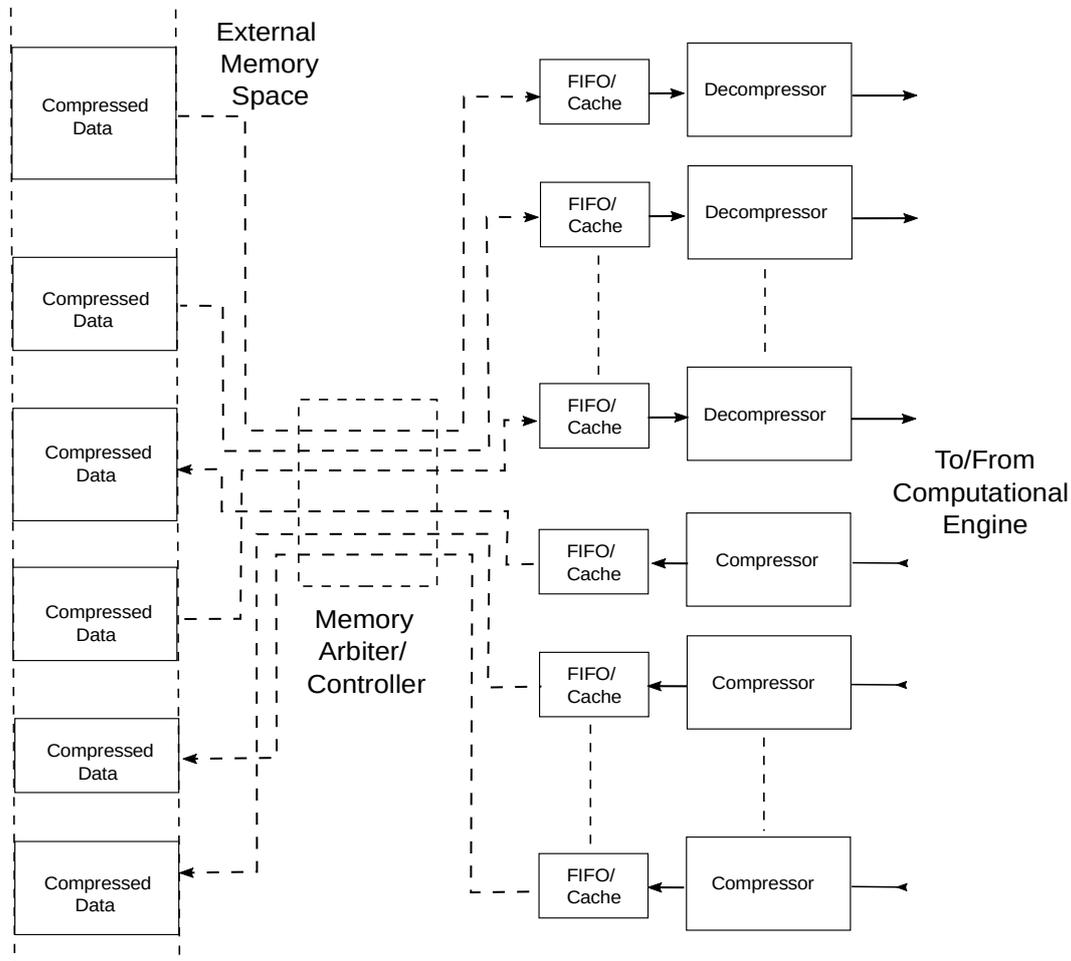

Figure 12 Streaming compressed data to/from a computational engine.

The compressors/decompressors shown in the figure can be considered inclusive of the interface to/from floating or fixed point as in section 4.3. or just coding pairs compressors/decompressors, depending on the processor they interface to.

Except for the reduced bandwidth and storage space, this is analogous to memory interface that fetches multiple streams of data to feed computational structures such as tensor cores and systolic arrays. The only difference is that the FIFOs and the memory will contain compressed data. As we have seen in section 4.4., it is possible to have random access to the compressed data, adding flexibility to the system.

The overhead in resources represented by the compressor/decompressors should be easily offset by the advantage of the bandwidth reduction: computation is cheaper than bandwidth.

## 4.7. An Inference Example: Weight Compression and Sharing

Computation in CNNs and LLMs (and the transformer architecture they are based on) is based in stages named layers. Given a particular architecture, the behaviour of each layer is dictated by the weights. CNNs and especially LLMs require a huge number of weights and, therefore, weight compression is particularly beneficial during inference as it reduces both the weight storage and the associated bandwidth.

In Llama2 7B there are ~7 billion parameters as bfloat numbers => ~14 Gbytes. So, in order to produce even a single token with this language model, we need to load in at least 14 Gbytes from memory (realistically DRAM as this is too large for on chip memory). And this is for the weights only.

It's clear that the ~34% reduction provided by the method in section 2.1. is especially welcome: a very inexpensive way to increase DRAM size and bandwidth by ~1.5x (more in conjunction with quantization).

Unfortunately, if we want to produce tens or hundreds of token per second, things scale up quickly and it's even worse for larger models.

There is, however, a category of problems where things can be simplified substantially: consider the case where multiple instances of the same model need to be applied to different data. This could be, for example, multiple, identical CNN models being applied to different frames from different cameras. Or a group of identical video transformers doing the same. Or a possibly popular one: multiple instances of the same LLM simultaneously processing multiple, unrelated queries (Fig.13).

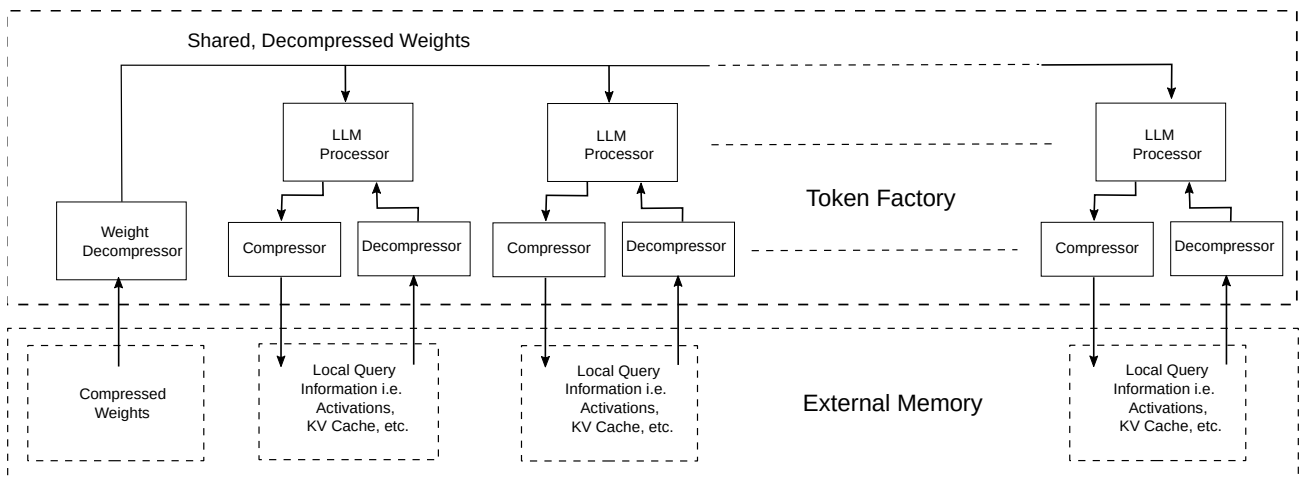

Figure 13 Token factory with shared and compressed weights.

This works well because, if we have multiple processors running the same LLM in lockstep on their private copy of the data, they will be requiring the same weights at the same time. The weight fetching and decompression infrastructure can then be shared amongst the processors.

The compressors/decompressors could be similar to Fig.12 while the weight decompressor will be similar to the decompression part only.

Since computation in LLMs like Llama2 is generally bandwidth bound, sharing the expensive weight fetching infrastructure is particularly advantageous: we can now run multiple relatively slow processors simultaneously, resulting in the production of hundreds of tokens per second albeit on separate queries.

For example, let's say that an acceptable level of performance is 10 tokens/s for each query for LLama2 7B. We need to design the LLM processor to be at least capable of that. Next we need to be able to load the weights at least 14*10 = 140 Gbytes/s, with compression 140*0.66 = ~92 Gbytes/s. We also need to take into account the storage for each query information but this normally much smaller than the weights and, in any case, it could be stored in a separate DRAM bank.

If we now run 10 of these LLM processors, we can produce 100 tokens/s requiring only ~92 Gbytes/s for the weight fetching instead of 100*14 Gbytes/s = 1.4 Tbytes/s without sharing and compression. This is just a back of the envelope type of calculation but it gives an idea of the advantage.

Note that all the LLM processors do not need to be perfectly synchronous: some lack of synchronization can be taken care with small buffers, allowing to distribute multiple LLM processors across multiple chips, all receiving the same weights from a single weight decompressor.

# 5. Conclusion

This paper has presented a simple framework that includes:

- Compressed, variable precision, variable range, user defined numerical data types that provide great flexibility in the computation of neural networks
- Bandwidth reduction which also means reduced storage and power in DRAMs. This result relies on statistical properties of the weights of CNNs and LLMs and it should be replicated in other cases
- A simple, programmable, low resources and high throughput hardware implementation of the above
- Benefits for research in weight and activations quantization that expands beyond the existing fixed formats

I really believe that adding such flexibility is both relatively easy and highly desirable in machine learning processors and GPUs and perhaps for other applications as well.

Given that, in GPUs, computation is cheaper than bandwidth, a software implement is also a possibility.